\documentclass{llncs}
\usepackage[misc]{ifsym} 
\usepackage{graphicx}
\usepackage{epstopdf}
\usepackage{float}
\usepackage{amssymb,amsmath}
\usepackage{bm}
\usepackage{multirow}
\usepackage[colorlinks,linkcolor=blue,anchorcolor=blue,citecolor=blue]{hyperref}
\begin{document}

\title{Direct Automated Quantitative Measurement of Spine via Cascade Amplifier Regression Network}
\titlerunning{Hamiltonian Mechanics}  
%
\author{Shumao Pang\inst{1} \and Stephanie Leung\inst{2} \and Ilanit Ben Nachum\inst{2} \and Qianjin Feng\inst{1}$^{\textrm{(\Letter)}}$ \and Shuo Li\inst{2}$^{\textrm{(\Letter)}}$}
\authorrunning{xxx et al.} 
%
\tocauthor{xxx, xxx, xxx, xxx,
xxx, xxx, and xxx}
\institute{Guangdong Provincial Key Laboratory of Medical Image Processing, School of Biomedical Engineering, Southern Medical University, Guangzhou, 510515, China \\
\email{qianjinfeng08@gmail.com}
\and Department of Medical Imaging, Western University, ON, Canada \\
Digital Imaging Group of London, ON, Canada\\
\email{slishuo@gmail.com}}

\maketitle              

\begin{abstract}
Automated quantitative measurement of the spine (i.e., multiple indices estimation of heights, widths, areas, and so on for the vertebral body and disc) is of the utmost importance in clinical spinal disease diagnoses, such as osteoporosis, intervertebral disc degeneration, and lumbar disc herniation, yet still an unprecedented challenge due to the variety of spine structure and the high dimensionality of indices to be estimated.
In this paper, we propose a novel cascade amplifier regression network (CARN), which includes the CARN architecture and local shape-constrained manifold regularization (LSCMR) loss function, to achieve accurate direct automated multiple indices estimation. The CARN architecture is composed of a cascade amplifier network (CAN) for expressive feature embedding and a linear regression model for multiple indices estimation.
The CAN consists of cascade amplifier units (AUs), which are used for selective feature reuse by stimulating effective feature and suppressing redundant feature during propagating feature map between adjacent layers, thus an expressive feature embedding is obtained.
During training, the LSCMR is utilized to alleviate overfitting and generate realistic estimation by learning the multiple indices distribution.
Experiments on MR images of 195 subjects show that the proposed CARN achieves impressive performance with mean absolute errors of 1.2496$\pm$1.0624 mm, 1.2887$\pm$1.0992 mm, and 1.2692$\pm$1.0811 mm for estimation of 15 heights of discs, 15 heights of vertebral bodies, and total indices respectively. The proposed method has great potential in clinical spinal disease diagnoses.
\end{abstract}
\vspace{-1.0cm}
\section{Introduction}
\begin{figure}[H]
\centering
\includegraphics[width=10cm]{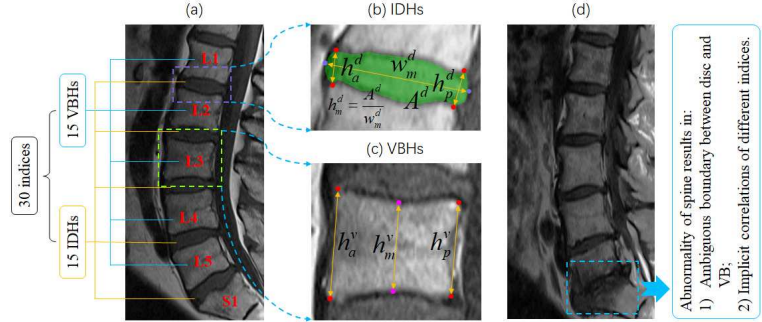}
\caption{(a) Illustration of 30 indices to be estimated; (b) Three heights for each disc (i.e., anterior IDH \emph{h}$_a^d$, middle IDH \emph{h}$_m^d$, and posterior IDH \emph{h}$_p^d$); (c) Three heights for each vertebral body (i.e., anterior VBH \emph{h}$_a^v$, middle VBH \emph{h}$_m^v$, and posterior VBH \emph{h}$_p^v$), where \emph{A}$^d$ denotes the disc area; (d) Ambiguous boundary between disc and VB and implicit correlations of different indices due to spinal abnormality.}
\label{fig:height}
\end{figure}

\vspace{-0.3cm}
The quantitative measurement of the spine (i.e., multiple indices estimation of heights, widths, areas, and so on for the vertebral body and disc) plays a significant role in clinical spinal disease diagnoses, such as osteoporosis, intervertebral disc degeneration, and lumbar disc herniation.
Specifically, the vertebral body height (VBH) and intervertebral disc height (IDH) (as shown in Fig. \ref{fig:height}) are the most valuable indices for the quantitative measurement of the spine.
The VBHs are correlated with the bone strength, which is of great significance to the vertebral fracture risk assessment for the osteoporotic patients \cite{McCloskey2012,Taton2014}.
Furthermore, the IDH reduction is associated with the intervertebral disc degeneration \cite{Jarman2014,Salamat2016} and lumbar disc herniation \cite{Tunset2013}.
\vspace{-0.7cm} 
\setlength{\abovecaptionskip}{-0.05cm}   
\setlength{\belowcaptionskip}{-0.8cm} 
Automated quantitative measurement of the spine is of significant clinical importance because it is reliable, time-saving, reproducible, and has higher consistency compared with manual quantitative measurement, which is usually obtained by manually detecting landmarks of the intervertebral disc (ID) and vertebral body (VB) from MR image \cite{Tunset2013,Videman2014}.

Direct automated quantitative measurement of the spine is an exceedingly intractable task due to the following challenges:
1) The high dimensionality of estimated indices (as shown in Fig. \ref{fig:height}(a)), which leads to difficulty in expressive feature embedding for such complex regression problem.
2) The excessive ambiguity of the boundary between VB and ID for abnormal spine (as shown in Fig. \ref{fig:height}(d)), which increases intractability of expressive feature embedding.
3) Implicit correlations between different estimated indices (as shown in Fig. \ref{fig:height}(d), the heights of the abnormal disc and the heights of adjacent VB are correlated because disc abnormality leads to simultaneous changes of IDH and the adjacent VBH), which is difficult to be captured.
4) Insufficient labelled data (as shown in Fig. \ref{fig:height}(d)), which possibly results in overfitting.

In recent years, an increasing number of approaches emerged in the direct quantitative measurement of other organs (e.g., heart)\cite{Zhen2017,Xue2017}. Although these methods achieved promising performance in the quantification of the cardiac image, they are incapable of achieving quantitative measurement of the spine because they suffer from the following limitations. 1) Lack of expressive feature representation. Traditional convolutional neural network (CNN) \cite{DBLP:journals/corr/SimonyanZ14a} is incapable of generating an expressive feature for multiple indices estimation because CNN possibly loses effective feature due to the lack of an explicit structure for feature reuse. 2) Incapability of learning the estimated indices distribution, which will lead to unreasonable estimation and overfitting.

In this study, we propose a cascade amplifier regression network (CARN), which includes the CARN architecture and local shape-constrained manifold regularization (LSCMR) loss function, for quantitative measurement of the spine from MR images. The CARN architecture is comprised of a cascade amplifier network (CAN) for expressive feature embedding and a linear regression model for multiple indices estimation.
In CAN, amplifier unit (AU) is used for selective feature reuse between adjacent layers. As shown in Fig. \ref{fig:CAN} (b), the effective feature of the anterior layer is stimulated while the redundant feature is suppressed, thus generating the selected feature, which is reused in posterior layer by a concatenation operator.
CAN reuses multi-level features selectively for representing complex spine, thus an expressive feature embedding is obtained.
During training, the high dimensional indices can be embedded in a low dimensional manifold due to the correlations between these indices.
LSCMR is employed to restrict the output of the CARN to the target output manifold. As a result, the distribution of the estimated indices is close to the real distribution, which reduces the impact of outliers and alleviates overfitting. Combining the expressive feature embedding produced by CAN with LSCMR, a simple linear regression model, i.e., fully connected network, is sufficient to produce accurate estimation results.

The main contributions of the study are three-fold.
1) To the best of our knowledge, it is the first time to achieve automated quantitative measurement of the spine, which will provide a more reliable metric for the clinical diagnosis of spinal diseases.
2) The proposed CAN provides an expressive feature map for automated quantitative measurement of the spine.
3) Overfitting is alleviated by LSCMR, which utilizes the local shape of the target output manifold to restrict the estimated indices to being close to the manifold, thus a realistic estimation of indices is obtained.
\vspace{-0.5cm}
\section{Cascade Amplifier Regression Network}
\begin{figure}[H]
\centering
\includegraphics[width=10cm]{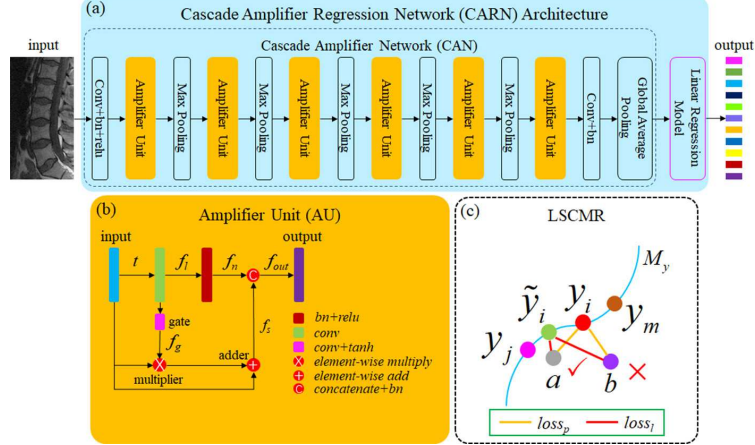}
\caption{(a) Overview of CARN architecture, including CAN for expressive feature embedding and a linear regression model for multiple indices estimation. (b) AU for selective feature reuse between adjacent layers. (c) LSCMR for obtaining realistic estimation and alleviating overfitting.}
\label{fig:CAN}
\end{figure}

\vspace{-0.3cm}
The CARN employs the CARN architecture and LSCMR loss function to achieve accurate quantitative measurement of the spine. The CARN architecture is composed of the CAN for expressive feature embedding and the linear regression model for multiple indices estimation. As shown in Fig. \ref{fig:CAN}, in CAN, AU is used for selective feature reuse between the adjacent layers by a gate, multiplier, adder and concatenate operator. In AU, the effective feature map is stimulated while the redundant feature map is suppressed. CAN provides expressive feature embedding via reusing multi-level features selectively.
The linear regression model in CARN is a fully connected network without non-linear activation. During training, overfitting is alleviated by LSCMR, which is employed to oblige the output of CARN to lie on the target output manifold expressed by local linear representation \cite{Pang2017}, i.e., a sample on the manifold can be approximately represented as a linear combination of several nearest neighbors from the manifold. Local linear representation captures the local shape of the manifold, therefore, the distribution of estimated indices is close to the real distribution and the indices estimated by CARN are realistic.
\vspace{-0.5cm}
\subsection{Mathematical Formulation}
\vspace{-0.3cm}
Automated quantitative measurement of the spine is described as a multi-output regression problem. Given a training dataset $T{\rm{ = }}\left\{ {{x_i},{y_i}} \right\}_{i = 1}^N$, we aim to train a multi-output regression model (i.e., the CARN) to learn the mapping $f:x \in {R^{h \times w}} \to y \in {R^d}$, where ${x_i}$ and ${y_i}$ denote the MR image and the corresponding multiple indices respectively, and $N$ is the number of training samples. CARN should learn an effective feature and a reliable regressor simultaneously.
\vspace{-0.7cm} 
\vspace{-0.5cm}
\subsection{CARN Architecture}
\vspace{-0.2cm}
The CARN architecture is comprised of the CAN for expressive feature embedding and the linear regression model for multiple indices estimation.

\textbf{CAN for Expressive Feature Embedding} The CAN consists of six AUs, two convolutional layers, five max pooling layers, and a global average pooling layer as shown in Fig. \ref{fig:CAN} (a). AU is designed for selective feature reuse between adjacent layers. During feature selection, the selected feature is obtained by amplifying the input feature of AU using an amplifier, whose amplification factor is learned automatically (details in Section Feature Selection Mechanism). The effective low-level feature is stimulated and concatenated by the high-level feature while the redundant low-level feature is suppressed. The selective feature reuse is achieved by CAN level by level; then the multi-level selective reused feature generates an expressive feature embedding.
The first convolutional layer with a $7\times7$ kernel size and stride of $2$ reduces the resolution of feature maps from $512\times256$ to $256\times128$, while the last convolutional layer with a $1\times1$ kernel size and stride of $1$ linearly combines the feature maps for information integration. The max pooling with a $2\times2$ kernel size and a stride of 2 is used to provide translation invariance to the internal representation. The global average pooling layer is utilized to reduce the dimensionality of feature maps.

The most crucial component of CAN is AU (as demonstrated in Fig. \ref{fig:CAN} (b)), which is composed of a gate for controlling information propagation between adjacent layers, a convolutional layer with a $3\times3$ kernel size and stride of $1$ for extracting a linear feature map, which is used to control the gate, a batch normalization layer with relu activation for producing non-linear feature map, a multiplier, an adder, and a concatenation operator with batch normalization for combining the selected feature map and non-linear feature map. The input $t$ of AU goes through a convolutional layer and produces the linear feature map $f_l\left( t \right) = w_l * t + b_l$ for guiding feature selection, where $w_l$ and $b_l$ are the convolution kernel weight and bias of the convolutional layer respectively, and $*$ is the convolutional operator. Then the $f_l\left( t \right)$ flows into two paths. One path consists of batch normalization and relu activation, which is analogous to the traditional CNN to generate non-linear feature map ${f_n}\left( t \right) = relu\left( {bn\left( {{f_l}\left( t \right)} \right)} \right)$, where $bn$ and $relu$ denote the batch normalization and relu activation respectively. The other path is a gate composed of a convolutional layer and tanh activation, which generates output ${f_g}\left( t \right) = tanh \left( {{w_g} * {f_l}\left( t \right) + {b_g}} \right)$, where $w_g$ and $b_g$ are the convolution kernel weight and bias in the gate respectively, for selecting feature map. The output of the gate flows into a multiplier followed by an adder, and generates the selected feature:
\setlength{\abovedisplayskip}{-0.5pt}
\setlength{\belowdisplayskip}{-0.5pt}
\begin{equation}
\begin{aligned}
  {f_s}\left( t \right) = t \odot f_g\left( t \right) + t = t \odot \left( {f_g\left( t \right) + 1} \right)
  \label{eq:four}
\end{aligned}
\end{equation}
where $\odot$ denotes the element-wise multiplication.
Finally, the $f_n$ and $f_s$ are concatenated along the channel axis and normalized by the batch normalization layer to generate a output feature map ${f_{out}}\left( t \right) = bn\left( {{f_n}\left( t \right) \oplus {f_s}\left( t \right)} \right)$, where $\oplus$ denotes the concatenation operator.

\textbf{Feature Selection Mechanism} In Eq. \ref{eq:four}, the value of each pixel in the selected feature map $f_s$ is obtained by multiplying an amplification factor with the corresponding value in the input feature map $t$. The amplification factor $[f_g(t)+1]$ ranges from 0 to 2; substantially, the selected feature map $f_s$ is equivalent to stimulating or suppressing the input feature map via an amplifier. When the amplification factor is less than 1, the input feature map is suppressed, vice versa. If the amplification factor is 1, the input feature map is directly propagated to the output, which is analogous to the denseNet \cite{Huang2017}.

\textbf{Linear Regression Model for Multiple Indices Estimation}
The linear regression model is a fully connected layer. The output of the linear regression model is: $f\left( {{x_i}} \right) = {w_o}h\left( {{x_i}} \right) + {b_o}$, where $h(x_i)$ is the output of the global average pooling (i.e., the feature embedding) as shown in Fig. \ref{fig:CAN} (a), and $w_o$ and $b_o$ are the weights matrix and bias of the linear regression respectively.
\vspace{-0.5cm}
\subsection{Local Shape-constrained Manifold Regularization Loss Function}
\vspace{-0.2cm}
The loss function is divided into two parts, including preliminary loss $loss_p$ and LSCMR loss $loss_m$. The preliminary loss is designed to minimize the distance between the estimation of indices and the ground truth, while the LSCMR loss is aimed at alleviating overfitting and generating realistic results by obliging the output of CARN to lie on the target output manifold using local linear representation. The total loss function is defined as follows:
\begin{equation}
{loss_{t}}\left( w \right) = {loss_p}\left( w \right) + {\lambda _l}{loss_l}\left( w \right)
\label{eq:nine}
\end{equation}
where the $\lambda _l$ is a scaling factor controlling the relative importance of the LSCMR loss.
The preliminary loss function is defined as follows:
\begin{equation}
  {loss_p}\left( w \right) = \frac{1}{{N \times d}}\sum\limits_{i = 1}^N {{{\left\| {{y_i} - f\left( {{x_i}} \right)} \right\|}_1}}  + {\lambda _p}\sum\limits_i {{{\left\| {{w_i}} \right\|}_2}}
  \label{eq:six}
\end{equation}
where the first term is the mean absolute error (MAE) of the regression model; the second term is the $l_2$ norm regularization for the trainable weight $w_i$ in CARN; $\lambda_p$ is a hyper-parameter.

By using only the preliminary loss function, unreasonable multiple indices estimation may be obtained because the estimated result is possible to be out of their real distribution. For instance, as shown in Fig. \ref{fig:CAN} (c), $y_i$, $y_j$, and $y_m$ are the target outputs of samples. The points $a$ and $b$ are two possible estimations of $y_i$. The distances between the two estimations (the points $a$ and $b$) and the target output $y_i$ are the same, ie., they have an identical preliminary loss. However, the loss of point $a$ should be smaller than the point $b$ as $a$ is much closer to the local shape of the output space than $b$. Hence, $a$ is a better estimation of $y_i$ than $b$.

LSCMR is proposed to achieve a realistic and accurate estimation of multiple indices.
Inspired by \cite{Liu2009}, ${y_i}$ lies on a manifold $M_y$ with an inherent dimension smaller than $d$ as the elements of $y_i$ are correlated. The manifold $M_y$ is spanned by ${\{y_i\}}_{i = 1}^N$. We introduce the local linear representation, i.e., a sample on manifold $M_y$ can be approximately represented as a linear combination of several nearest neighbors from $M_y$ \cite{Pang2017}. A sample $y_i$ on $M_y$ is locally linearly represented as:
\begin{equation}
\begin{aligned}
\begin{array}{l}
{y_i} = \sum\limits_{j = 1}^k {{y_j}{\alpha _j}}  + \varepsilon  \approx \sum\limits_{j = 1}^k {{y_j}{\alpha _j}}  = {{\tilde y}_i}\\
s.t.\left\| \varepsilon  \right\| < \tau ,\sum\limits_{j = 1}^k {{\alpha _j}}  = 1,{\alpha _j} \ge 0,{y_j} \in N({y_i})
\end{array}
\label{eq:seven}
\end{aligned}
\end{equation}
where $\varepsilon$ is the reconstruction error and $\tau$ is a small non-negative constant. $N({y_i})$ denotes the $k$-nearest neighbors of $y_i$ on $M_y$ and $\alpha _j$ is the reconstruction coefficient, which is calculated by LAE \cite{liu2010large}. As shown in Fig. \ref{fig:CAN} (c), ${\tilde y}_i$ is the local linear representation of $y_i$ using its $k$-nearest neighbors (here $k$ is equal to 2) $y_j$ and $y_m$. The local linear representation of $y_i$ reflects the local manifold shape. If the predicted indices is close to ${\tilde y}_i$, it will be near the manifold $M_y$. Therefore, the LSCMR loss is defined as:
\begin{equation}
los{s_l}\left( w \right) = \frac{1}{{N \times d}}\sum\limits_{i = 1}^N {{{\left\| {f\left( {{x_i}} \right) - {{\tilde y}_i}} \right\|}_1}}
\label{eq:eight}
\end{equation}
Using the $loss_l$, the prediction of $y_i$ is restricted to being close to the manifold $M_y$, thus a more realistic result is obtained (e.g., the model generate the point $a$ as the estimation of $y_i$ instead of point $b$ in Fig. \ref{fig:CAN} (c)).
\vspace{-0.5cm}
\section{Experimental Results}
\vspace{-0.2cm}
\textbf{Dataset} The dataset consists of 195 midsagittal spine MR images from 195 patients. The pixel spacings range from 0.4688 mm/pixel to 0.7813 mm/pixel. Images are resampled to 0.4688 mm/pixel and the ground truth values are obtained manually in this space. In our experiments, two landmarks, i.e., the left-top corner of the L1 VB and the left-bottom corner of the L5 VB, are manually marked for each image to provide reference for ROI cropping, in which five VBs, including L1, L2, L3, L4 and L5, and five IDs under them are enclosed. The cropped images are resized to $512 \times 256$.

\textbf{Experimental Configurations} The network is implemented by Tensorflow. Four group experiments under different configurations, including CARN-${loss_p}$, CNN-${loss_p}$, CNN-${loss_t}$, and CARN-${loss_t}$, are used to validate the effectiveness of our proposed method. In CNN-${loss_p}$ and CNN-${loss_t}$, AU is replaced with a traditional convolutional layer, in which the output feature channels are the same as AU; the -$loss_p$ and -$loss_t$ denote the loss function defined in Eq. \ref{eq:six} and Eq. \ref{eq:nine} respectively used in the model.

\textbf{Overall Performance} As shown in the last column of Table \ref{table:results}, the proposed CARN achieves low error for automated quantitative measurement of the spine, with MAE of 1.2496$\pm$1.0624 mm, 1.2887$\pm$1.0992 mm, and 1.2692$\pm$1.0811 mm for IDHs, VBHs, and total indices respectively. These errors are small referring to the maximums of IDHs (20.9203 mm) and VBHs (36.7140 mm) in our dataset.

\textbf{CAN and LSCMR Effectiveness}
Combining CAN and LSCMR, the performance improved by 2.44\%, 1.16\%, and 1.80\% for IDHs, VBHs, and total indices estimation respectively, which is clearly demonstrated by comparing the third and last columns of Table \ref{table:results}.
Using CAN without LSCMR, the MAE decreased by 0.21\%, 0.49\%, and 0.36\% for IDHs, VBHs, and total indices estimation respectively, as shown in the second and third columns of Table \ref{table:results}. These results indicate that the CAN improves the performance for total indices estimation, especially for VBHs. This results from the fact that CAN generates expressive feature embedding although pathological changes in the disc can reduce the intensity of the adjacent VB and lead to ambiguity in the boundary.
Using LSCMR without CAN, the performance improved by 2.14\%, 1.03\% for IDHs, and total indices estimation respectively, as shown in the third and fourth columns of Table \ref{table:results}.

LSCMR alleviates overfitting as shown in Table \ref{table:results}, in which CARN-$loss_t$ and CNN-$loss_t$ have high training errors (0.8591 mm vs 0.5024 mm, 0.9059 mm vs 0.5224 mm) but low test errors (1.2692 mm vs 1.2878 mm, 1.2791 mm vs 1.2924 mm) for total indices estimation compared with CARN-$loss_p$ and CNN-$loss_p$.
\vspace{-0.4cm} 
\begin{table}
\caption{Performance of CARN in terms of MAE under different configurations for IDH (mm), VBH (mm), and total indices (mm) estimation. MAE is illustrated in each cell. Best results are bolded for each row.}
\vspace{-0.4cm}
\begin{center}
\begin{tabular}{c|c|c|c|c|c}
\hline
\hline
\multicolumn{2}{c|}{Method} & CARN-$loss_p$ & CNN-$loss_p$ & CNN-$loss_t$ & CARN-$loss_t$ \\
\hline
\multirow{2}*{IDH} & train & 0.4633$\pm$0.4706 & 0.4920$\pm$0.4574 & 0.8689$\pm$0.7417 & 0.8265$\pm$0.7012 \\
                   & test  & 1.2782$\pm$1.1173 & 1.2809$\pm$1.1172 & 1.2535$\pm$1.0754 & \textbf{1.2496$\pm$1.0624} \\
\hline
\multirow{2}*{VBH} & train & 0.5414$\pm$0.5846 & 0.5528$\pm$0.5615 & 0.9429$\pm$0.8383 & 0.8916$\pm$0.8004 \\
                   & test  & 1.2974$\pm$1.0922 & 1.3038$\pm$1.1154 & 1.3047$\pm$1.1215 & \textbf{1.2887$\pm$1.0992} \\
\hline
\multirow{2}*{Total} & train & 0.5024$\pm$0.5321 & 0.5224$\pm$0.5130 & 0.9059$\pm$0.7923 & 0.8591$\pm$0.7531 \\
                     & test  & 1.2878$\pm$1.1049 & 1.2924$\pm$1.1163 & 1.2791$\pm$1.0990 & \textbf{1.2692$\pm$1.0811} \\
\hline
\hline
\end{tabular}
\end{center}
\label{table:results}
\end{table}
\vspace{-1.5cm}
\section{Conclusions}
\vspace{-0.3cm}
We have proposed a multi-output regression network CARN for automated quantitative measurement of spine. By taking advantage of expressive feature extracted from CAN, and employing LSCMR for alleviating overfitting, CARN is capable of achieving promising accuracy for all indices estimation.
\\
\textbf{Acknowledgements.}This work was supported by China Scholarship Council (No. 201708440350), National Natural Science Foundation of China (No. U1501256), and Science and Technology Project of Guangdong Province (No. 2015B010131011).
\vspace{-0.4cm}
%
%
%
\bibliographystyle{splncs}
\bibliography{reference}
\end{document}